# 8590 | Modelling of Terrain Deformation by a Grouser Wheel for Lunar Rover Simulation

Junnosuke Kamohara[a,*,†], Vinicius E. Ares[b,†], James Hurrell[a], Keisuke Takehana[a], Antoine Richard[c], Shreya Santra[a], Kentaro Uno[a], Eric Rohmer[b], Kazuya Yoshida[a]

[a] The Space Robotics Lab. (SRL) in Department of Aerospace Engineering, Tohoku University, Sendai, Miyagi, Japan

[b] The Advanced Robotics Lab. (AdRoLab) in School of Electrical and Computer Engineering, State University of Campinas, Brazil

[c] The Space Robotics Research Group at the Interdisciplinary Research Center for Security reliability and Trust (SnT), University of Luxembourg, Luxembourg

† These authors contributed equally.
* Corresponding author: kamohara.junnosuke.t6@dc.tohoku.ac.jp

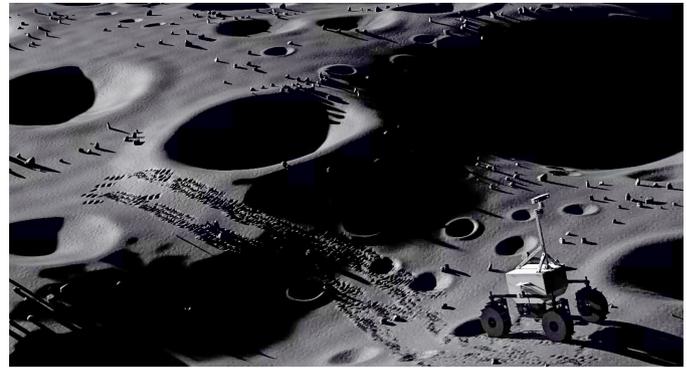

Fig. 1. Simulation of a rover with generated wheel traces.

## ABSTRACT

Simulation of vehicle motion in planetary environments is challenging. This is due to the modeling of complex terrain, optical conditions, and terrain-aware vehicle dynamics. One of the critical issues of typical simulators is that they assume terrain is a rigid body, which limits their ability to render wheel traces and compute the wheel-terrain interactions. This prevents, for example, the use of wheel traces as landmarks for localization, as well as the accurate simulation of motion. In the context of lunar regolith, the surface is not rigid but granular. As such, there are differences in the rover's motion, such as sinkage and slippage, and a clear wheel trace left behind the rover, compared to that on a rigid terrain. This study presents a novel approach to integrating a terramechanics-aware terrain deformation engine to simulate a realistic wheel trace in a digital lunar environment. By leveraging the Discrete Element Method simulation results alongside experimental single-wheel test data, we construct a regression model to derive deformation height as a function of contact normal force. The region of interest in a height map is retrieved from the wheel poses. The elevation values of corresponding pixels are subsequently modified using contact normal forces and the regression model. Finally, we apply the determined elevation change to each mesh vertex to render wheel traces during runtime. The deformation engine is integrated into our ongoing development of a lunar simulator based on NVIDIA's Omniverse IsaacSim. We hypothesize that our work will be crucial to testing perception and downstream navigation systems under conditions similar to outdoor or terrestrial fields. A demonstration video is available here: https://www.youtube.com/watch?v=TpzD0h-5hv4

## 1. Introduction

*1.1. Background*

The Lunar and Martian surfaces have seen a resurgence of interest as a focus of recent space explorations. Recently, India, Japan, and the United States have accomplished landings on the lunar surface. In the upcoming years, space agencies and private companies are looking toward long-term resource utilization on the lunar surface.

Robotic systems play a vital role in these future missions as they can reach hazardous areas that humans cannot or yet explore cost-effectively. These systems are expected to autonomously complete various tasks such as exploration, sample return, and construction. For robotic systems to operate effectively on a planetary surface, their hardware and software must be thoroughly tested in advance. A typical software test consists of two stages: the first is conducted using simulators, and the second in analog environments.

The simulation ranges from dedicated simulation of mechanical parts to end-to-end mission simulation. Particularly in robotics, a simulator can play a role in multiple areas: perception, navigation, mobility, radiation simulation, and mission planning. Today, there are several robotic simulators, such as Gazebo, CoppeliaSim (Rohmer et al. 2013), Unity, Unreal Engine, and IsaacSim. These simulators have the fundamental capability to simulate multiple sensors and vehicle dynamics with built-in rigid-body physics engines. However, one of the critical limitations of typical simulators is that they assume terrain as a rigid body. Therefore, they cannot represent wheel traces and compute wheel-terrain interactions (i.e., terramechanics). In reality, lunar regolith is not rigid but granular, so the granular material properties must be simulated.

In this work, we extend our previous work on the lunar simulator, OmniLRS (Richard et al., 2024), and focus on providing a terramechanics-aware visual representation of wheel trace in lunar environments. The simulation of wheel traces is essential in validating vision-based systems, such as Visual Inertial SLAM, with similar visual observation as the real-world sensory data. By incorporating a terramechanics-aware visual representation of wheel traces into the simulator, we provide a more accurate and valuable tool for the development and testing of lunar exploration rovers.

*1.2. Significance of simulating wheel traces*

On uneven terrains such as the lunar and Martian surfaces, wheel slippage during rover traversal can occur, potentially resulting in the rover becoming stuck. To minimize slippage and design an effective vehicle control and body, it is imperative to study terramechanics, which considers soil deformation.

Terramechanics is the field of study that addresses the interaction between soil and machines, focusing primarily on the characteristics traversing over unpaved, deformable soil. Research in this field has been progressing since the 1950s when Bekker developed fundamental models of wheel mechanics (Bekker, 1956). Subsequently, traction models were developed and refined by Wong et al. (Wong et al., 1967, 2008).

In recent years, research in the field of terramechanics has focused on the trace formed by wheels passing over soil. These traces can reflect the state of the wheel when they are created. In particular, the slip condition, which is crucial for rover control, can be estimated from the spacing of these traces. Saku et al. use machine learning to estimate slip conditions from the trace formed during single-wheel tests (Saku et al., 2019). Similarly, Reina et al. have proposed a vision-based method for estimating sideslip angles by observing the wheel traces, useful for the rover's localization and control systems (Reina et al., 2008). Therefore, the development of simulators that consider soil deformation plays a critical role in estimating the driving state.

## 2. Related Works

In the field of planetary robotics, there are usually two types of simulators: one focusing more on sensor simulation (Allan et al., 2019; Crues et al., 2022; Bingham et al., 2023; Richard and Kamohara et al., 2024; Müller et al., 2021; Pieczyński et al., 2023), and the other on accurate physics computation (Tasora et al., 2016; Krenn and Hirzinger, 2009; Zhou et al., 2023; Wang et al., 2022; Serban et al., 2023). The former is usually built upon a game engine, featuring realistic illumination and shadow rendering thanks to GPU-accelerated rendering. The latter typically supports a dedicated physics engine for wheel-terrain interactions or terramechanics.

### 2.1. Lunar ground simulators

As of today, there are various lunar ground simulators developed using different rendering engines. Allan et al. describe a Gazebo-based simulator aimed at simulating NASA Ames' VIPER missions. Although it incorporates many crucial features for lunar exploration, such as large-scale terrain management, realistic surface illumination, realistic shadows, and the opposition effect, the rendering of wheel traces is still limited to a primitive method. In their work, the wheel traces are simulated by manipulating the normal map, one of the texture images. The same approach is used by Pieczyński et al. to realize wheel traces in UnrealEngin4 (Pieczyński et al., 2023). However, these approaches have limitations in simulating other types of sensors such as LiDAR and Radar, which require mesh geometry.

Digital Lunar Exploration Sites (DLES) by Crues et al., is another simulator developed by NASA Johnson Space Center to simulate landing and rover exploration in the lunar south pole. It thoroughly explains the procedure of pre-processing digital elevation maps and managing large-scale terrain. Furthermore, it presents a database to efficiently store and query the coordinates and sizes of craters and boulders. High-resolution terrain models, along with surface details, rocks, and craters from DLES, are incorporated into the DLES Unreal Simulation Tool (DUST) by Bingham et al. to simulate the lunar south pole. However, these works still lack terramechanics consideration, particularly realistic wheel traces for vision-based systems.

OmniLRS is a robotic simulator for lunar analog environments, based on Nvidia's Omniverse Isaac-Sim. It provides two types of virtual environments: the Lunalab, a lunar analogue at the University of Luxembourg, and the Lunaryard, a procedurally generated lunar terrain. With its synthetic data generation pipeline, it demonstrated sim-to-real transfer from the virtual environment to the real-world lab environment in rock segmentation tasks. However, it still lacks features as a comprehensive robotics simulator, such as visual-inertial odometry, SLAM, and navigation. Additionally, it lacks wheel traces simulation, which limits its ability to replicate real-world scenarios and test perception and navigation systems in such settings.

### 2.2. Terramechanics in simulators

Another essential simulation tool is the simulator dedicated to terramechanics. This type of simulator focuses on accurate terrain-wheel modeling and the computation of the forces and torques applied to the wheels.

The Discrete Element Method (DEM), originally formalized by Cundall and Strack (1979), is a numerical approach used to study the interaction between particulate materials and solid objects, such as rover wheels. DEM provides detailed insights into the complex dynamics of wheel-soil interaction. Its utility in terramechanics is well-documented in the literature, such as in O'Sullivan (2011). Additionally, Lichtenheldt et al. (2016) highlight its relevance in planetary exploration vehicle design and its ability to simulate terrain deformation to form wheel traces, also known as rutting. Wasfy et al. (2014) explored the coupling of multibody dynamics with DEM for vehicle mobility on granular terrains, highlighting the detailed physical interactions. However, it does not address real-time simulation involving control and navigation in the loop because of the high computational cost.

Because of DEM's heavy computational demand, less accurate, but more feasible empirical terramechanics models are on the rise. Chrono (Tasora et al.) is an open-source C++ framework designed for physics-based modeling and simulation. The library simulates various multibody systems, such as wheeled and tracked vehicles on deformable terrains, robots, mechatronic systems, compliant mechanisms, and fluid-solid interactions. Particularly, Chrono supports deformable terrain, modeled by a soil contact model (SCM) (Krenn and Hirzinger), and allows computing force between terrain and the arbitrary contact surface, and deforming mesh in real-time. Serban et al. improved the computational efficiency of Chrono SCM by minimizing the number of ray-casting via the moving patch approach and parallelizing ray-casting. However, its usage is limited to a few-meter ordered field, and there is no study on perception and navigation using the deformable terrain.

There is also an effort to integrate terramechanics into common robotics simulators. MarsSim (Zhou et al.) is a simulator built upon Gazebo that supports rendering of the Martian landscape and includes a custom-built terramechanics plugin. The plugin is based on a semi-empirical model, both from the theory and experimental data, capable of obtaining drawbar pull, wheel slip, and wheel sinkage. In one of their experiments, they tested the transition from soft sand to hard bedrock. Unfortunately, their work does not consider the rendering of wheel traces, thus its application in vision systems is still limited.

In this work, we focus on integrating soil deformation in the previously developed photorealistic lunar simulator based on Nvidia IsaacSim. Our contribution is twofold: first, we model deformation depth and geometry in a data-driven manner using

DEM simulation. Second, we integrate this deformation engine into our previous lunar simulator, OmniLRS, allowing for rover operation with ROS1 or ROS2 while rendering wheel traces, as shown in Fig. 1. Computationally intensive DEM simulations provide accurate deformation data that is used to build a linear regression model, that can then run in real-time. We evaluate the quality of the simulated wheel traces by comparing them with images from hardware experiments.

## 3. Method

In this work, we propose a method to simulate wheel traces in a virtual lunar environment. Our system is composed of the following modules: the DEM simulation, a wheel trace model, real-time mesh deformation in IsaacSim. Figure 2 provides an overview of our system, with each component described in detail in the following sections.

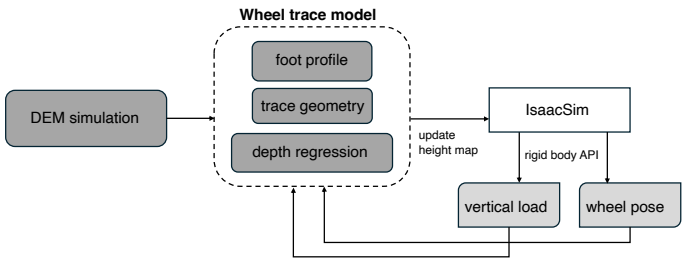

*Fig. 2. The figure shows the entire architecture of the proposed method. It takes vertical load and wheel pose to update the height map during simulation.*

### 3.1. Discrete Element Method

A data-driven approach is employed to model soil deformation using the Discrete Element Method (DEM) simulation, validated against experimental results from single-wheel testing. We utilize DEM to retrieve trace information, including trace geometry and resulting depth values for vertical loads acting on the wheel.

We use the EDEM (Altair, 2024), DEM simulation software, to compute wheel trace dimensions from input vertical load. The DEM simulations replicate a single-wheel test (SWT) experiment, where a wheel is driven with controlled velocities in a sandbox under a given load and output the trace geometry. Inside the EDEM, SWT is performed for a set of vertical loads, and the resulting wheel traces are measured to build the wheel trace model that can be used in IsaacSim.

We use the material called Toyoura sand (Ozaki et al., 2023) which is known for its low cohesion value and a narrow size distribution with a similar average particle size to lunar regolith. To reduce simulation time, particles size is scaled by a factor of ten and modelled as single spheres with the scaled average particle size. As for the contact model between particles and objects, we employ a standard Hertz-Mindilin model (Coetzee, 2017) with a Type-C rolling contact model (Ai et al., 2011). The simulation parameters are calibrated following our previous study (Hurrell et al., 2023) that uses the Angle of Repose (AoR) experiment to calibrate the parameters.

### 3.2. Wheel Trace Geometry

To simulate wheel traces, we implement (1) a foot profile generator, (2) $xz$ and $yz$ cross-section generators, and (3) a trace depth regression model. As for the coordinate frame, we use three types: the local coordinate of the robot's link, the global coordinate, and the pixel coordinate of the reference height map. The local coordinate follows the FLU convention, in which the x-axis faces robot's forward motion direction and z-axis faces upward.

#### 3.2.1. Foot profile

First, the foot profile is created based on the geometry and dimensions of the links in contact with the surface (these links are wheels in our case). From now on, we call these links in contact with the surface *target links*. The foot profile is a set of two-dimensional points, defined in each target link's local coordinate, occupied by their projections. In the case of a wheeled robot, this profile is a set of points inside rectangles which are the projections of the wheels. We assume the width and height of the rectangular profile are in the same dimensions as the width ($w$) and diameter of the wheels ($2r$). These sets of points inside the foot profile, expressed in local coordinates, are indicated in Eq. 1. In the end, the derived points are transformed from their local coordinates to global coordinates and then to pixel coordinates. These transformed points indicate the pixel positions where deformation is applied.

$$FP = \left\{(x,y) \mid -r \leq x \leq r, -\frac{w}{2} \leq y \leq \frac{w}{2}\right\} \quad (1)$$

#### 3.2.2. xz, yz cross-section

Once the foot profile is computed, we define the distribution of the deformation profile. This profile tells us the scattering distribution of the wheel trace $\{z = f(x,y) \mid (x,y) \in FP\}$. In our work, we assume the trace distribution is a combination of two distributions. Based on the results from DEM simulations, we propose using a trigonometric function and a trapezoidal curve for $h(x)$ and $g(y)$, as they approximate the DEM results well. For the $xz$ and $yz$ cross-sections, we use the functions in Eq. 2 and Eq. 3.

$$h(x) = \cos(n\pi x); n = \frac{N}{2\pi r(1-s)} \quad (2)$$

$$g(y) = \begin{cases} \tan(\theta)\frac{2y}{w} - \tan(\theta) & \left(1 - \frac{1}{\tan(\theta)} < \frac{2y}{w} \leq 1\right) \\ -1 & \left(-1 + \frac{1}{\tan(\theta)} \leq \frac{2y}{w} \leq 1 - \frac{1}{\tan(\theta)}\right) \\ -\tan(\theta)\frac{2y}{w} - \tan(\theta) & \left(-1 \leq \frac{2y}{w} \leq -1 + \frac{1}{\tan(\theta)}\right) \end{cases} \quad (3)$$

Here, $n$ and $\theta$ are the frequency of the wave computed from the number of grousers ($N$) and the slip ratio ($s$), and the internal angle of repose of the soil material. In the end, the deformation profile is computed as shown in Eq. 4 by combining Eq. 2 and Eq. 3. $A(F_z)$ and $\mu(F_z)$ are computed from the vertical load - depth regression model discussed in the following section.

$$z = g(y) \cdot \left(A(F_z) \cdot h(x) + \mu(F_z)\right) \quad (4)$$

#### 3.2.3. Depth regression model

The last component in the trace geometry generator is the depth regression model. This regression model, as a function of vertical load, determines the depth value of the amplitude $A(F_z)$ and the mean depth $\mu(F_z)$ of the wheel trace. We use pairs of vertical load and deformation profiles retrieved from DEM simulations to find the slope and *y* intercept of the linear regression model.

### 3.3. Integration in IsaacSim

The final step is to integrate our deformation engine in IsaacSim to simulate the robot locomotion with wheel traces. We manipulate the reference digital elevation map and update the terrain mesh during runtime to create wheel traces in Isaac.

#### 3.3.1. Mesh deformation

As discussed in previous sections, the deformation engine requires the vertical load applied on the terrain and the pixel positions of the target links. We use Isaac's rigid body API (*RigidPrimView*) to compute the vertical load applied on the terrain and the pose of the wheels. Taking these values as input, our deformation engine computes a set of pixel indices to apply deformation and the elevation change. Then, the digital elevation map is updated, and the *z* position of the mesh vertices is changed accordingly.

In practice, we only update the position of the mesh vertices, but not the collision. This is because IsaacSim needs to encode the mesh into a specific data structure so that its physics engine can understand it to compute collisions. However, this encoding process takes a lot of time and pauses the simulation. Therefore, accurate physics computation is not within the scope of our current work.

#### 3.3.2. The vehicle dynamics

For the simulation, we use a multi-body robot model of the four-wheeled custom rover (Rodríguez-Martínez et al., 2024). The model is exported from a CAD model to the Universal Robot Description Format (URDF) and the Universal Scene Description (USD), which is compatible with IsaacSim. The resulting model includes a set of rigid bodies, joints (fixed, revolute, and prismatic), and a collider that represents mesh collision.

When simulating the robot, the robot state, such as pose, velocity, and acceleration, is updated by the PhysX5 rigid body engine, taking actuator commands as inputs. These actuator commands consist of the angular velocity of the four driving motors and the steering angle of the two front steering motors. These low-level actuator commands are computed using the Ackermann steering model (Simionescu and Beale, 2002), with inputs of forward velocity and steering angle. The control inputs are then applied to the robot using a PD controller implemented in Isaac.

## 4. Implementation

Primary applications of wheel trace simulation include vision-based tasks such as Visual-Inertial-SLAM and vision-based navigation. Wheel traces are crucial for simulating outdoor terrains in high fidelity because they are common landmarks seen in the field. We demonstrate that our simulator can simulate a robot's locomotion with visually realistic traces. As a qualitative evaluation, we compare the images from our simulator with those from outdoor experiments.

### 4.1. DEM simulation settings

Inside the DEM simulation, we used the wheel of the Rashid rover (Almaeeni, Els and Almarzooqi, 2024), designed for the Emirates lunar mission to explore the Atlas crater. The wheel has a diameter of 200 mm, a width of 80 mm, and 14 grousers of 20 mm in length. The virtual SWT sandbox measures 650 mm in length, 140 mm in width, and 120 mm in depth. Particles are modeled as spheres with a homogeneous size distribution. Due to the sub-millimeter size of Toyoura sand particles and the high computational cost, a scaling factor of ten was used, resulting in particles with a 1 mm radius. For terrain parameters, we used those from Hurrell et al. Table 1 shows the list of terrain parameters used in DEM simulation.

Table 1
List of terramechanics parameters used in DEM simulation.

| Wheel and particle geometry | Value |
|---|---|
| Wheel diameter | 200 mm |
| Wheel width | 80 mm |
| Grouser length | 20 mm |
| Grouser count | 14 |
| Sandbox length | 650 mm |
| Sandbox width | 140 mm |
| Sandbox depth | 120 mm |
| Particle geometry | sphere |
| Particle diameter | 2 mm |
| Particle volume | 4.189 mm$^3$ |
| **Geometry discretization and particle counts** | |
| Wheel facets | $15.6 \times 10^3$ |
| Wheel nodes | 7,802 |
| Sandbed box facets | 10 |
| Sandbed box nodes | 8 |
| Particle count | $1.38 \times 10^6$ |
| Contacts count total | $2.948 \times 10^6$ |
|     Particle-particle contact count | $2.901 \times 10^6$ |
|     Wheel-particle contact count | 1,582 |
|     Sandbox-particle contact count | 45,200 |
| **Material properties and contact model** | |
| Particle material | Toyoura sand |
|     Poisson ratio | 0.25 |
|     Shear modulus | $2.0 \times 10^7$ Pa |
|     Density | 2,650 kg/m$^3$ |
| Wheel material | Aluminum |
|     Poisson ratio | 0.33 |
|     Shear modulus | $2.71 \times 10^{10}$ Pa |
|     Density | 2,810 kg/m$^3$ |
| Particle-particle interaction | |
|     Coefficient of restitution | 0.40 |
|     Coefficient of static friction | 0.65 |
|     Coefficient of rolling friction | 0.15 |
| Wheel-particle interaction | |
|     Coefficient of restitution | 0.40 |
|     Coefficient of static friction | 0.60 |
|     Coefficient of rolling friction | 0.40 |
| Contact Model | Hertz-Mindlin |
| Coefficient of rolling stiffness | 3 |
| Rolling viscous damping ratio | 0.3 |

The SWT DEM simulation begins by filling the sandbox with particles to form the sand-bed and allowing the particles to settle. Next, the depth, bulk density, and level surface are checked, removing excess particles if necessary to ensure they match the desired conditions. The wheel is then inserted with the required load and allowed to settle for one second. The wheel is driven with the set horizontal and rotational velocities, and the simulation is allowed to process until the total time is reached and the travel is finished. After the simulation, a cross-section is created in the plane *y*=0 to reveal the wave pattern of the wheel trace. Finally, the heights of the trough and peak of the wave shape are measured relative to the original height of the sand-bed.

*4.2. Simulation in IsaacSim*

For the simulation, we used a procedurally generated lunaryard as explained in Richard et al. We chose the synthetic lunar terrain because it provides more control over the size and resolution of the terrain. The terrain was generated with a size of 20 m × 20 m and a resolution of 1.5 cm/texel. As for the texture, we used sand-like materials as well as finer gravel materials from Polyhaven and Nvidia's Base Material collection.

To simulate the dynamics of rovers, we must have a realistic model of robot. As mentioned in the previous section, we used the EX1 rover (Rodríguez-Martínez et al., 2024), a four-wheeled rover with steering developed by our group. The wheels of the EX1 rover are scaled to match the dimensions of those on the Rashid rover. To observe the wheel traces from the rover's point of view, two cameras were attached to the front and back of the rover. Since the actual EX1 hardware uses the Intel D435 camera (Keselman, 2017), we used the same camera intrinsic parameters.

## 5. Results and Discussion

To integrate wheel traces in IsaacSim, we create the wheel trace model from DEM simulation. Then, we leverage this model to compute how much terrain mesh deforms and render traces during runtime. First, we present the results from the DEM simulation and introduce a regression model that computes deformation depth. Finally, we compare simulated and actual images from the outdoor field test.

*5.1. DEM simulation*

Fig.3 presents $xz$ and $yz$ cross-sections of wheel traces from DEM simulation. To measure the dimensions of wheel traces, we used a ruler tool available inside EDEM. When looking at $yz$ cross-section, it resembles the shape of a trapezoid, making Eq. 3 a reasonable assumption. From several profiles in the $xz$ plane, the heights of the trough and peak of wheel traces were measured for a set of loads, as presented in Table 2.

Table 2
*Measured values of trough, crest, mean, and amplitude of the wheel trace in the SWT simulation for each applied vertical load.*

| Vertical Load [N] | Trough height [mm] | Crest height [mm] | Mean height [mm] | Amplitude [mm] |
|---|---|---|---|---|
| 3.0 | -6.5 | 1.5 | -2.50 | 8.0 |
| 5.0 | -8.0 | 0.5 | -3.75 | 8.5 |
| 8.6 | -9.5 | -1.0 | -5.25 | 8.5 |
| 12.9 | -12.0 | -3.0 | -7.50 | 9.0 |
| 17.9 | -13.5 | -4.5 | -9.00 | 9.0 |

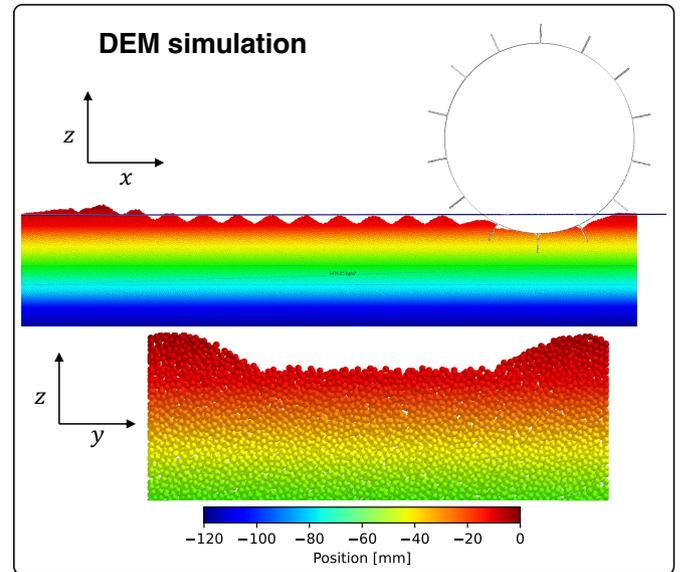

*Fig. 3. The figure shows xz and yz cross-section of the wheel trace in the EDEM simulation.*

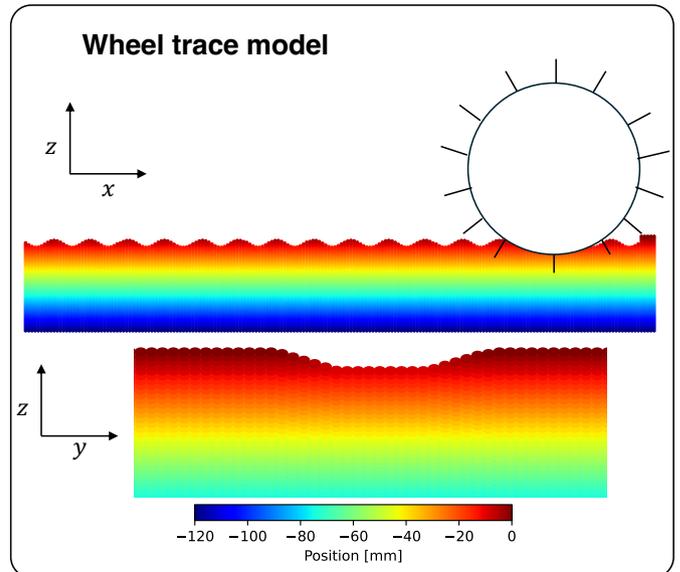

*Fig. 4. The figure shows the wheel trace modeled by our proposed method. Same as Fig.3, the top image indicates xz cross-section and the bottom image represents yz cross-section.*

The values of average height and amplitude are used to build linear regression models to predict average height and amplitude as functions of the vertical load. Eq. 5 and Eq. 6 show the computed linear regression models for average height and amplitude. The fitted model for average height yielded a mean absolute error of 0.16 mm and a mean percentage error of 3.5%. In the case of amplitude, it had a mean absolute error of 0.14 mm and a mean percentage error of 1.6%. The regression models can then be used in Eq. 4 to render wheel traces in the next section.

$$\mu(F_z) = -1.32 - 0.458 Fz \quad (5)$$

$$A(F_z) = 7.98 + 0.066 Fz \quad (6)$$

## 5.2. Qualitative evaluation of wheel trace rendering

Inside IsaacSim, we drove the EX1 rover on flat terrain and recorded the rendering with wheel traces. The rover was operated inside a 20 m × 20 m procedurally generated lunar terrain under an illumination source with 3,000 intensity. Fig. 5 shows the rendering of wheel traces and images from the outdoor field test at a sand dune in Tottori, Japan. The image comparison suggests that the grouser-shaped wheel traces are well simulated in our simulator. However, we still have difficulty rendering the trace with more surface roughness and sharpness. This is because our current texture is in 4,000-pixel resolution, which is not high enough to represent higher surface roughness. The texture problem will be addressed in future work.

Another advantage of our method is its scalability to other types of wheeled robots, as shown in Fig. 6. This is because our method has control over the dimensions of the foot profile, which is adjustable through configuration files. In the future, we will support other types of robots, such as quadrupedal robots, by considering arbitrary contact foot geometry.

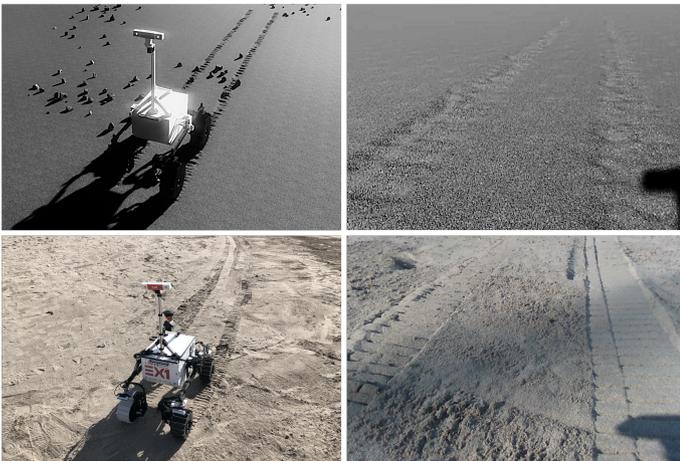

*Fig. 5. Comparison between the rendering of wheel traces in IsaacSim and the images from the outdoor field test. Top two images show rendering in simulation and the bottom two show images from the outdoor field test.*

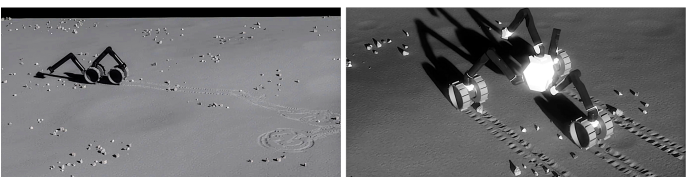

*Fig. 6. Rendering of wheel traces created by other types of wheeled robots.*

## 5.3. Computation time and memory complexity of simulation

In the end, we analyzed the mesh update time and VRAM consumption for different terrain sizes and resolutions. The terrain sizes chosen were 10m × 10m, 20m × 20m, and 40m × 40m, with resolutions of 0.1m/texel, 0.2m/texel, 0.25m/texel, and 0.5m/texel. To compare mesh update time and GPU memory consumption against the mesh dimensions, we used the product of mesh size and the inverse of squared resolution (i.e., texels) as mesh dimension metrics. Figure 7 shows the plot of computation time and GPU usage on the vertical axis and texels on the horizontal axis. The figure suggests that both mesh update time and GPU memory consumption linearly increase. According to the figure, we can assume that a reasonable mesh size is up to one million texels (20m × 20m at 0.02m/texel), which results in 12.9ms. Furthermore, the GPU usage decreases as the number of texels increases. This is happening as the update of the mesh is happening at the CPU level, as the number of texels increases, the time it takes to update the mesh gets increasingly longer reducing the overall GPU usage.

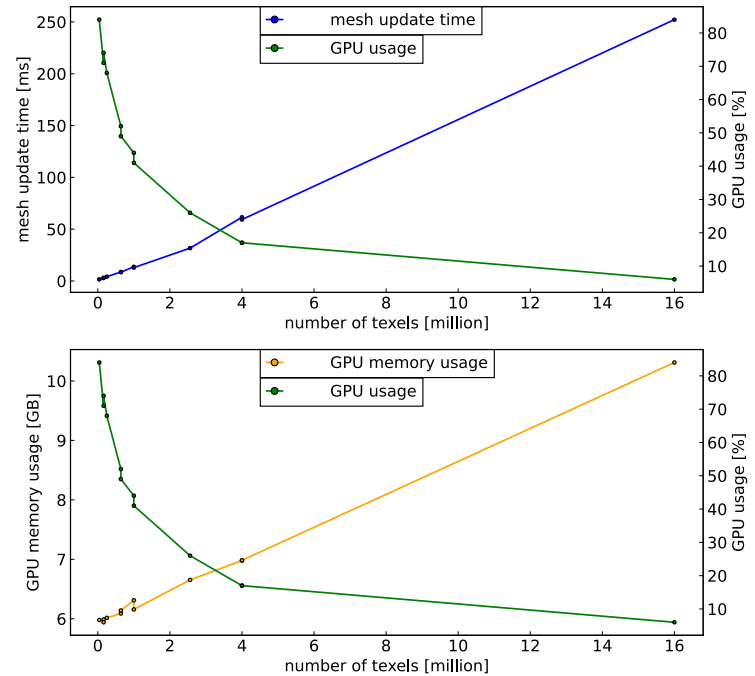

*Fig. 7. The diagram shows the plot of mesh update time, GPU memory consumption, and GPU usage against the number of texels.*

## 5.4. Future work

As for future work, we will include the wheel trace in large-scale, kilometer-order terrain and simulate long-range traversal. Then, we will study if visual SLAM can leverage wheel traces as additional landmarks to increase the accuracy of localization.

In addition, the mesh deformation time is upper bounded by the streaming of mesh vertex points done on the CPU. A faster method to stream vertices will reduce the computation time, making our method scalable to finer and larger terrains.

## 6. Conclusion

This study proposed a method to simulate wheel traces for lunar rovers using a developed terramechanics-aware deformation engine. By utilizing DEM simulations, we developed a regression model to compute deformation height based on contact normal forces and integrated it into IsaacSim. Our method overcomes the limitations of rigid terrain assumptions in existing simulators and realizes real-time wheel traces in the simulation loop.

Our approach is scalable to different types of wheeled robots, rendering wheel traces with specified contact surface geometry and dimensions. Future work will extend our simulator to be able to handle large-scale terrain and simulate long-range traverses with wheel traces. This feature allows us to test visual inertial SLAM with and without wheel traces, and study if SLAM can leverage wheel traces to increase localization accuracy. Furthermore, visual analysis of the wheel traces left by the rover can provide valuable insights into the vehicle's state and performance during its traversal, which will eventually aid in mission planning.

## 7. Acknowledgements

The authors would like to acknowledge Kenta Sawa, Masahiro Uda, and Kohta Naoki for kindly providing the images recorded from the EX1 rover at the Tottori sand dunes. This work was supported by JST Moonshot R&D Program, Grant Number JPMJMS223B. This study was financed in part by the Coordenação de Aperfeiçoamento de Pessoal de Nível Superior (CAPES) - PrInt Program- Brazil (Finance Code 001 - Process no. 1041304P), of which author V.E. Ares is a beneficiary.

## 8. Declaration of competing interest

The authors declare that they have no known competing financial interests or personal relationships that could have appeared to influence the work reported in this paper.